%% file: example_paper.tex
\documentclass{article}

\usepackage[preprint]{icml2026}
\usepackage{microtype}
\usepackage{graphicx}
\graphicspath{{figures/}}
\usepackage{subcaption}
\usepackage{booktabs}
\usepackage{amsmath}
\usepackage[T1]{fontenc}
\usepackage[utf8]{inputenc}
\usepackage{hyperref}
\usepackage{inconsolata}

\icmltitlerunning{Temporal Linguistic Emergence in Diffusion Language Models}

\begin{document}

\twocolumn[
  \icmltitle{Measuring Temporal Linguistic Emergence in Diffusion Language Models}

  \begin{icmlauthorlist}
    \icmlauthor{Harry Lu}{yyy}
  \end{icmlauthorlist}

  \icmlaffiliation{yyy}{School of Mathematics, University of Minnesota, Minneapolis MN, United States}
  \icmlcorrespondingauthor{Harry Lu}{lu000661@umn.edu}
  \icmlkeywords{diffusion language models, mechanistic interpretability, probing}

  \vskip 0.3in
]

\printAffiliationsAndNotice{}

\begin{abstract}
\input{sections/0-Abstract}
\end{abstract}

\input{sections/1-Intro}
\input{sections/2-RelatedWorks}
\input{sections/3-Methods}
\input{sections/4-Experiments}

\bibliographystyle{icml2026}
\bibliography{example_paper}

\clearpage
\appendix
\input{sections/5-Appendix}

\end{document}

%% file: sections/0-Abstract.tex
Diffusion language models expose an explicit denoising trajectory, making it possible to ask when different kinds of information become measurable during generation. We study three independent 32-step runs of \texttt{GSAI-ML/LLaDA-8B-Base} on masked WikiText-103 text, each with 1{,}000 probe-training sequences and 200 held-out evaluation sequences. From saved trajectories, we derive four temporal measurements: token commitment; linear recoverability of part-of-speech (POS), coarse semantic category, and token identity; confidence and entropy dynamics; and sensitivity under mid-trajectory re-masking. Across seeds, the same ordering recurs: content categories stabilize earlier than function-heavy categories, POS and coarse semantic labels remain substantially more linearly recoverable than exact lexical identity under our probe setup, uncertainty remains higher for tokens that ultimately resolve incorrectly even though late confidence becomes less calibrated, and perturbation sensitivity peaks in the middle of the trajectory. A direct/collateral decomposition shows that this peak is overwhelmingly local to the perturbed positions themselves. In this LLaDA+WikiText setting, denoising time is therefore a useful analysis axis: under our measurements, coarse labels are recovered earlier and more robustly than lexical identity, trajectory-level uncertainty tracks eventual correctness, and mid-trajectory states are the most intervention-sensitive.

%% file: sections/1-Intro.tex
\section{Introduction}
Autoregressive language models reveal little about \emph{when} a prediction becomes stable: generation advances left-to-right, and intermediate states are not naturally aligned to a fixed denoising clock. Diffusion language models, by contrast, iteratively refine a masked sequence and therefore expose a temporal axis that can be analyzed directly \citep{li2022diffusionlm,nie2025llada}. That temporal structure is especially attractive for interpretability. If representations and predictions evolve over denoising steps, we can ask whether coarse linguistic labels, lexical identity, certainty, and intervention sensitivity follow the same schedule or distinct ones.

We analyze three independent full 32-step runs of \texttt{GSAI-ML/LLaDA-8B-Base} on masked WikiText-103 text \citep{merity2016pointer}. Our analysis is intentionally lightweight: it operates only on saved trajectories, uses simple linear probes in the spirit of prior probing work \citep{hewitt2019structural,liu-etal-2019-linguistic,rogers2020primer}, and pairs descriptive summaries with a single intervention-sensitivity curve. Even under these constraints, the temporal structure is far from uniform.

We do not claim a universal law of diffusion-language generation from three runs. Instead, we ask whether four temporal measurements follow a consistent ordering across seeds: prediction stability, linearly recoverable linguistic information, self-reported certainty, and sensitivity to mid-trajectory re-masking. They do, but only in a measurement-aware sense. Under our probes, coarse labels are more recoverable than exact lexical identity throughout denoising, uncertainty remains informative about eventual correctness even late in the trajectory, and a non-initial re-masking-sensitive window becomes disproportionately important for the final outcome. Because the lexical probe uses a compact, partially unseen label space, we treat these comparisons ordinally rather than as normalized effect sizes.

Our contributions are threefold: a lightweight trajectory-first framework for commitment, recoverability, certainty, and intervention sensitivity; empirical evidence, in one LLaDA+WikiText setting across three seeds, for a stable temporal ordering in which coarse labels are more recoverable than lexical identity under our probe setup and trajectory-level uncertainty tracks eventual correctness despite calibration drift; and a direct/collateral perturbation analysis showing that the strongest re-masking-sensitive window is non-initial and mostly local to the perturbed positions.

%% file: sections/2-RelatedWorks.tex
\section{Related Work}
Diffusion language modeling has been driven mainly by controllable generation, masked-LM initialization, scaling, and noise-schedule design \citep{li2022diffusionlm,he2022diffusionbert,ye2023diffusion,nie2025llada,vonrutte2025scaling}. Recent work also pushes DLMs toward AR-scale adaptation and hybrid causal decoding, including AR-to-diffusion conversion in DiffuLLaMA, non-Markovian trajectory conditioning in CaDDi, and causal/masked hybrids such as ARMD and CARD \citep{gong2024scaling,zhang2025caddi,karami2026armd,ruan2026card}. Trajectory-level work has also begun using denoising dynamics for downstream signals such as hallucination detection and generative stability \citep{hemmat2026tdgnet,gautam2026energy}, but these studies do not directly ask when linguistic structure becomes available during denoising. We focus on that temporal question.

Our commitment analysis is also related to stabilization and early-exit work outside diffusion, where intermediate-layer confidence is used to decide when further computation is unnecessary \citep{xin2020deebert,elhoushi2024layerskip}. That literature treats stability mainly as an efficiency signal; we instead use a fixed denoising clock to compare when different kinds of linguistic information become linearly decodable, prediction-stable, and causally important.

The analysis also connects to probing and representation analysis \citep{alain2016understanding,conneau-etal-2018-cram,belinkov-glass-2019-analysis,hewitt2019structural,liu-etal-2019-linguistic,tenney-etal-2019-bert,de-vries-etal-2020-whats,rogers2020primer}. Prior work uses simple probes to study what contextual encoders represent, often along network depth; we keep that probe family lightweight and move the analysis axis to denoising time. In this sense, our paper is closer in spirit to diffusion-trajectory interpretability in vision and planning-style diffusion, where temporal stages already serve as meaningful units of analysis \citep{tinaz2025emergence,chen2026chain}.

%% file: sections/3-Methods.tex
\section{Methods}
We treat each full denoising run as a token-by-step table defined over masked positions. For every masked token, the logger stores the predicted token, top-1 confidence, entropy, and final-layer hidden state at every denoising step. This representation lets us analyze the same trajectory from four complementary views: stabilization, recoverability, certainty, and intervention sensitivity.

\paragraph{Measurements.}
Let $M_r$ denote the masked positions in record $r$, let $\hat{y}^{(t)}_{r,i}$ be the prediction at step $t$ for masked position $i \in M_r$, let $h^{(t)}_{r,i}$ be the corresponding hidden state, and let $q^{(t)}_{r,i}$ and $H^{(t)}_{r,i}$ be the stored top-1 confidence and entropy. We introduce the four measurements one by one. First, token commitment is the earliest step after which a prediction no longer changes:
\begin{align}
c_{r,i} &= \min \{ t : \hat{y}^{(t')}_{r,i} = \hat{y}^{(t)}_{r,i}, \forall t' \ge t \}.
\end{align}
Second, for a probe target $g \in \{\text{POS}, \text{semantic}, \text{token}\}$, temporal recoverability is the average linear-probe accuracy over masked positions:
\begin{align}
A_g(t) &= \frac{1}{N}\sum_{r,i} \mathbf{1}\!\left[\arg\max W_g h^{(t)}_{r,i} = \ell^g_{r,i}\right],
\end{align}
where $N=\sum_r |M_r|$. Third, we summarize certainty at step $t$ by the mean top-1 confidence and entropy:
\begin{align}
\bar{q}(t) &= \frac{1}{N}\sum_{r,i} q^{(t)}_{r,i}, \qquad
\bar{H}(t) = \frac{1}{N}\sum_{r,i} H^{(t)}_{r,i}.
\end{align}
Finally, intervention sensitivity is the drop in final accuracy after perturbing the trajectory at step $t$:
\begin{align}
\Delta(t) &= \mathrm{Acc}_{\text{base}} - \mathrm{Acc}_{\text{perturb at } t}.
\end{align}
Throughout, we interpret this as sensitivity under a specific re-masking intervention rather than as a fully identified causal mechanism. For the direct/collateral follow-up, we additionally evaluate this drop separately on the re-masked positions themselves and on the untouched masked positions, yielding $\Delta_{\text{direct}}(t)$ and $\Delta_{\text{collateral}}(t)$. For the probe plots, we reuse per-record outputs to estimate bootstrap uncertainty.

\begin{figure*}[t]
  \centering
  \begin{subfigure}[t]{0.49\textwidth}
    \centering
    \includegraphics[width=\linewidth]{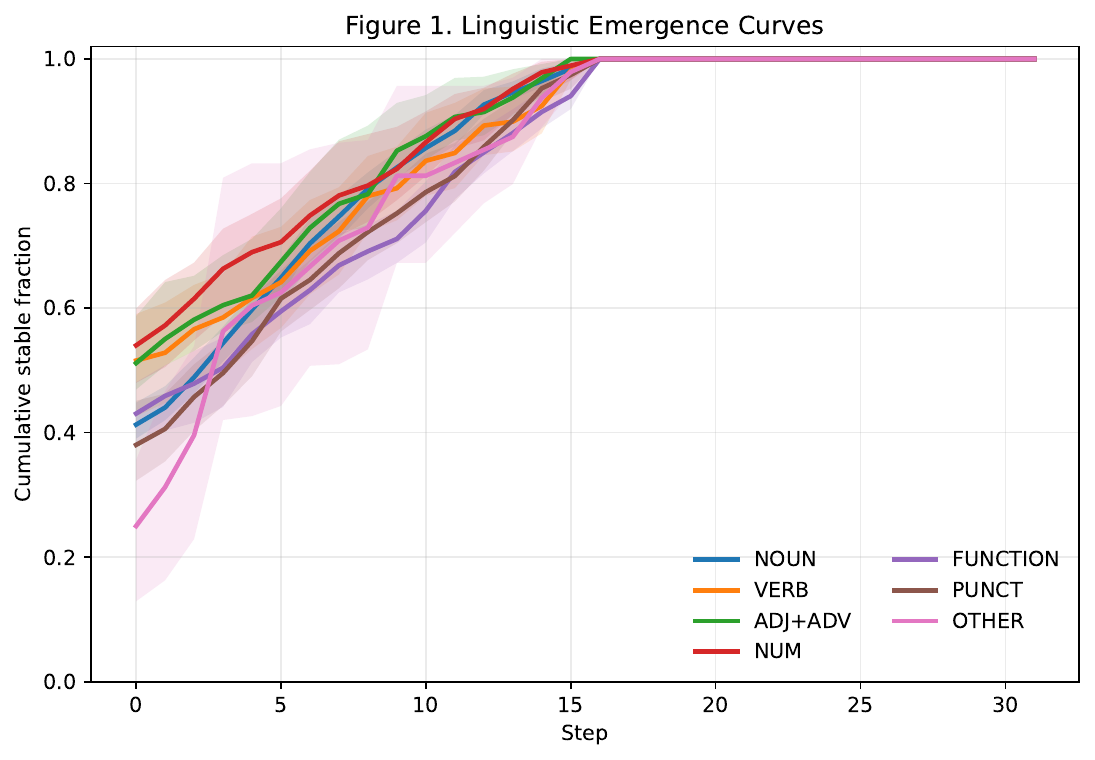}
    \caption{Commitment accumulates earlier for numbers and content words than for function-heavy categories.}
  \end{subfigure}\hfill
  \begin{subfigure}[t]{0.49\textwidth}
    \centering
    \includegraphics[width=\linewidth]{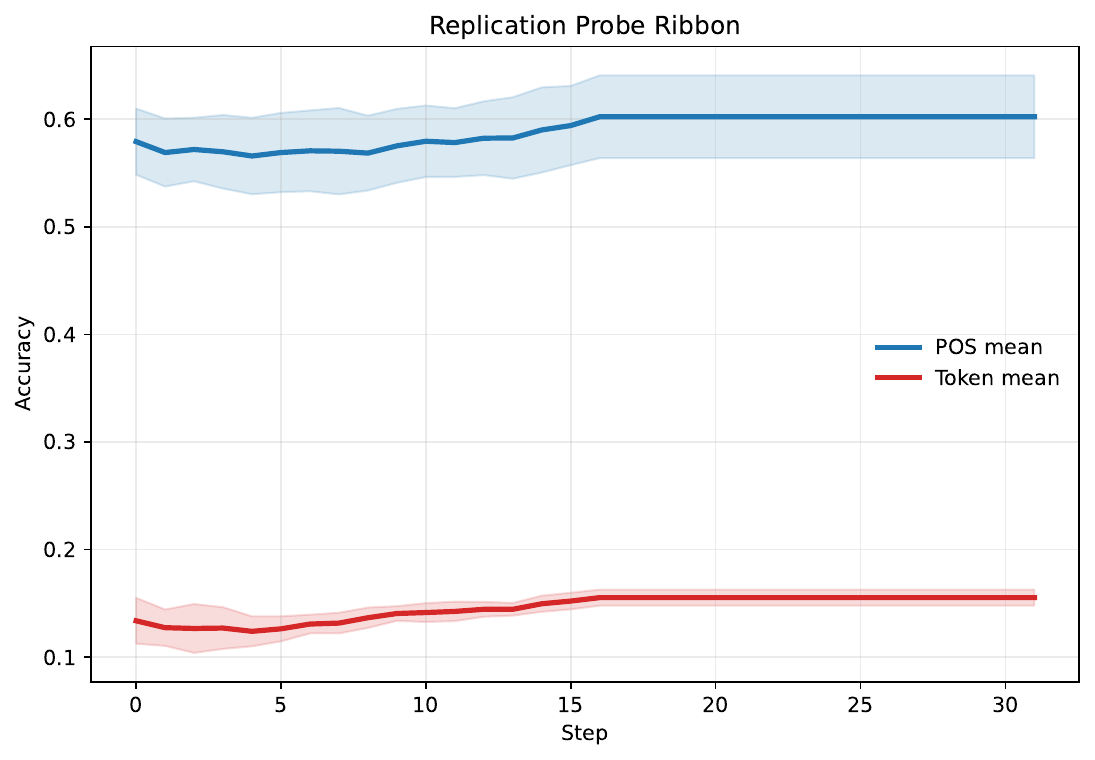}
    \caption{Across three seeds, coarse labels remain much more linearly recoverable than exact lexical identity.}
  \end{subfigure}

  \vspace{0.3em}

  \begin{subfigure}[t]{0.49\textwidth}
    \centering
    \includegraphics[width=\linewidth]{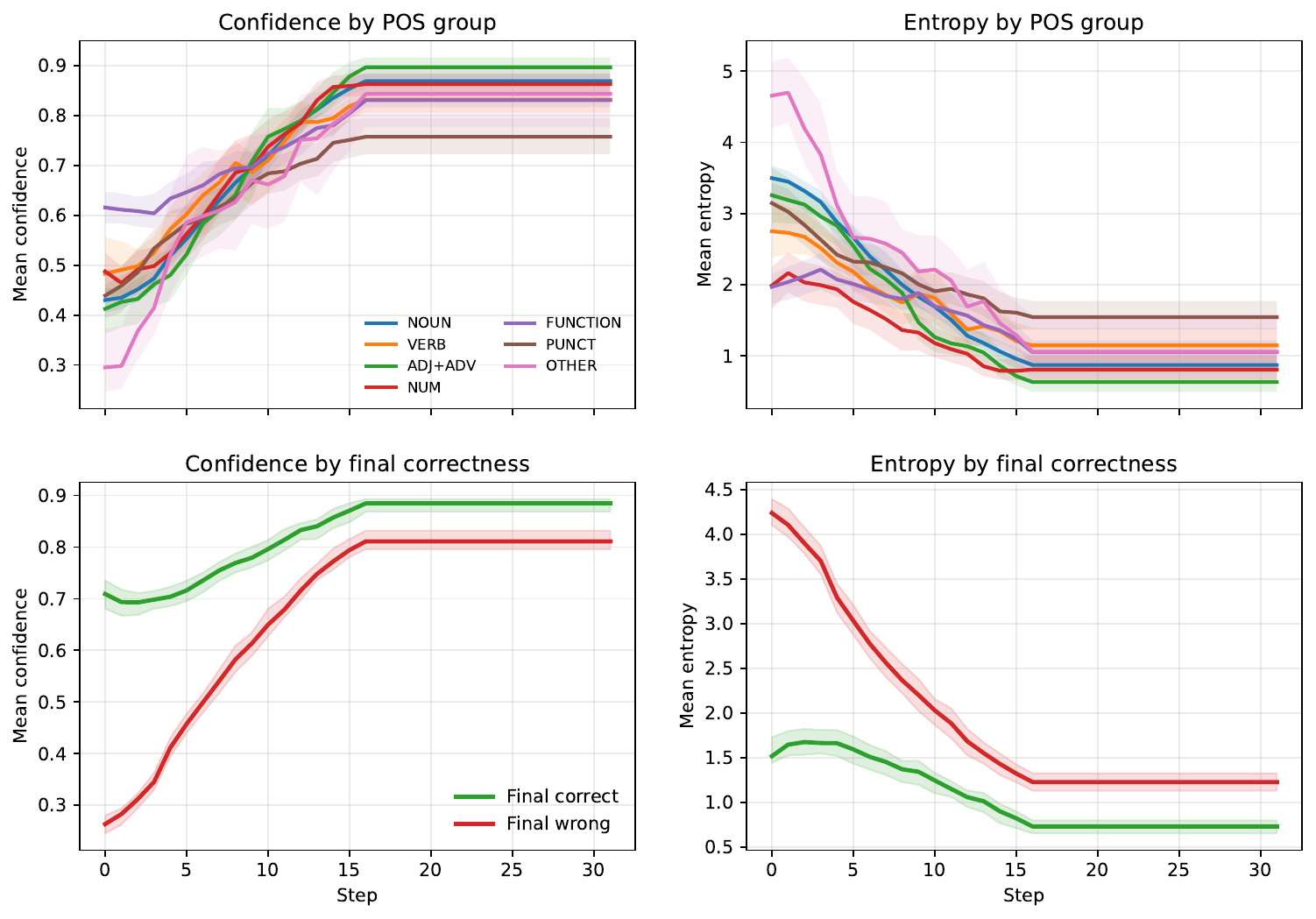}
    \caption{Trajectory-level uncertainty separates eventually correct from eventually wrong tokens late in denoising.}
  \end{subfigure}\hfill
  \begin{subfigure}[t]{0.49\textwidth}
    \centering
    \includegraphics[width=\linewidth]{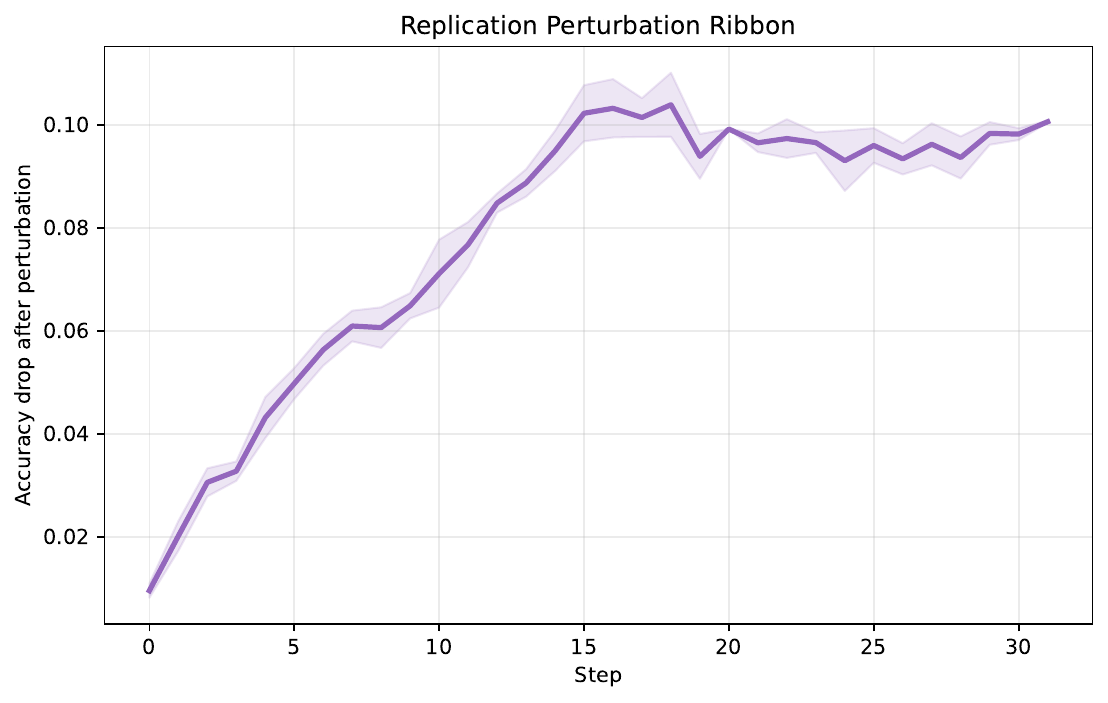}
    \caption{Across three seeds, perturbation sensitivity peaks in the middle of the trajectory rather than at step 0.}
  \end{subfigure}
  \caption{Temporal recoverability, uncertainty, and sensitivity summary for the 200-example evaluation split. Panels (b,d) are the main replicated results, panel (c) gives the focal uncertainty analysis, and panel (a) provides supporting commitment dynamics. Panels (a,c) show the representative seed-42 run, and panels (b,d) report mean $\pm$ one standard deviation over three seeds.}
  \label{fig:main}
\end{figure*}

For grouped commitment curves, let $z_{r,i}$ be the coarse POS group of masked position $(r,i)$ and let $N_k=\sum_{r,i}\mathbf{1}[z_{r,i}=k]$. We then report the empirical commitment CDF
\begin{align}
F_k(t) = \frac{1}{N_k}\sum_{r,i}\mathbf{1}[z_{r,i}=k]\mathbf{1}[c_{r,i}\le t],
\end{align}
which makes Figure~\ref{fig:main}(a) interpretable as a family of group-conditioned stabilization curves rather than only as mean commitment summaries.

\paragraph{Experimental setup and aggregation.}
We evaluate \texttt{GSAI-ML/LLaDA-8B-Base} on masked WikiText-103 trajectories over three independent 32-step runs and report either cross-seed mean $\pm$ standard deviation or a representative seed-42 trajectory when the qualitative pattern is visually stable. Full sampling, masking, and logging details are deferred to Appendix~A.

%% file: sections/4-Experiments.tex
\section{Results}
\paragraph{Coarse labels are more recoverable than lexical identity.}
Figure~\ref{fig:main}(b) aggregates the probe curves over three seeds. POS starts at 57.9\%\,$\pm$\,3.1 and ends at 60.2\%\,$\pm$\,3.8, coarse semantic labels rise from 59.8\% to 62.3\%, and token top-1 moves only from 13.4\%\,$\pm$\,2.1 to 15.5\%\,$\pm$\,0.7. Because the token probe is evaluated in a compact, partially unseen lexical label space (3{,}428 classes on average; 33.4\% of evaluation targets unseen during probe training), we interpret the POS/token gap ordinally rather than as a normalized effect size and complement top-1 with retrieval-style metrics (Appendix Table~\ref{tab:quant-summary-followup}). Even with that caveat, the ordering is stable: no step in any seed brings token identity close to POS or semantic recoverability, and the final POS--token gap is 44.7 points.

Aggregate lexical top-1 therefore understates what the probe can recover, but it does not erase the coarse-before-lexical pattern. Final token retrieval reaches 29.4\% top-5, 34.4\% top-10, and 0.218 MRR overall; on seen lexical targets, final top-1/top-5/top-10/MRR improve to 23.4/44.1/51.7/0.327, versus essentially zero on unseen targets. A per-step probe ablation still keeps the POS--token gap positive at every step in every run, never below 34.5 points. Under this probe setup, coarse labels are therefore consistently more linearly recoverable than exact lexical identity throughout denoising.
The same ordering survives compact follow-ups: on WikiText-2, final POS/semantic/token recoverability is 57.2/61.2/18.4; under a 16-step schedule it is 63.9/66.0/15.1; and with per-step rather than shared probes, final POS/token accuracies are 58.6 and 14.6, with a minimum POS--token gap of 34.5 points.

\paragraph{Trajectory-level uncertainty tracks eventual correctness.}
Figure~\ref{fig:main}(c) shows that the uncertainty split is already visible in the representative seed-42 trajectory, and the endpoint summaries replicate across seeds. Eventually correct tokens end at 0.877 confidence versus 0.819 for eventually wrong tokens, and at 0.774 entropy versus 1.173. The denoising clock therefore does more than track representational emergence: it also separates trajectories that converge to the right basin from those that do not.

\begin{figure}[t]
  \centering
  \includegraphics[width=\columnwidth]{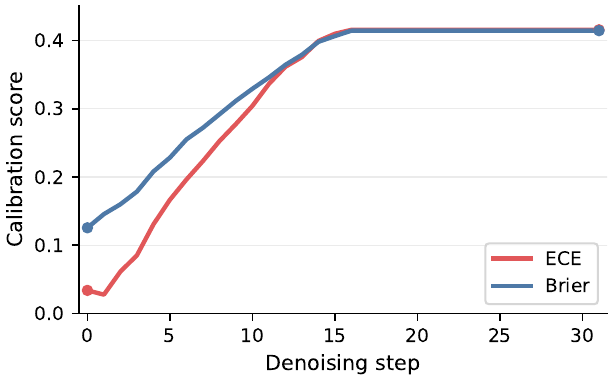}
  \caption{Calibration drift on the representative seed-42 run. ECE and Brier rise through the middle of denoising and plateau by about step 16, showing that later confidence is more separative but less calibrated.}
  \label{fig:uncertainty-calibration}
\end{figure}

Figure~\ref{fig:uncertainty-calibration} isolates the calibration drift behind this effect. On the representative run, ECE rises from 0.034 at step 0 to 0.415 at the final plateau, and the Brier score rises from 0.126 to 0.414, with both curves flattening around step 16 (Appendix Table~\ref{tab:quant-summary-followup}). Later confidence is therefore more discriminative yet less calibrated, which makes uncertainty a substantive interpretability result rather than just a cosmetic confidence trace.

\paragraph{Mid-trajectory perturbations are strongest and mostly local.}
Figure~\ref{fig:main}(d) shows the mean final accuracy drop after re-masking 20\% of tokens at step $t$ and resuming denoising, aggregated over three seeds. The curve rises from roughly one point at step 0 to a peak drop of about 10--11 points, with peak steps between 15 and 18. The most damage-sensitive window therefore appears after partial structure has formed rather than at initialization.

A direct-versus-collateral decomposition shows that this peak is overwhelmingly local. At the three-seed mean peak, the total drop is 10.96 points, of which 10.92 come from the re-masked positions and only 0.03 from untouched masked positions, so 99.7\% of the peak effect is direct. Collateral effects are not absent, but they are concentrated much earlier in denoising: at step 2, the mean drop is 3.13 points, with only 0.16 direct and 2.96 collateral. Selector-restricted controls preserve the same non-initial peak, which strengthens the interpretation that this is a robust re-masking-sensitive window rather than a brittle artifact of one perturbation subset (Appendix Tables~\ref{tab:quant-summary-perturb} and~\ref{tab:quant-summary-followup}).
Varying the perturbation design changes the amplitude more than the location of this effect: 10/20/40\% re-masking peaks at 7.1/11.2/19.9 points in the same late-middle region, and selector-restricted controls show the same non-initial sensitivity window, although changing the initial mask ratio shifts the denoising clock itself.

\paragraph{Commitment timing adds supporting temporal structure.}
Figure~\ref{fig:main}(a) shows that commitment timing has stable category structure across seeds: numbers commit earliest on average (mean 3.37 steps, range 3.23--3.59), nouns and verbs are intermediate (4.07 and 4.46), and function-heavy or punctuation tokens are latest (4.92 and 4.97). We use commitment as a stabilization descriptor rather than a correctness proxy, since the representative seed-42 run is non-monotonic in correctness: tokens that commit at step 0 are correct 62.2\% of the time, mid-trajectory commitments (steps 5--9) drop to 30.9\%, and very late commitments (step 10 onward) recover to 49.3\% (Figure~\ref{fig:commitment-correctness}; Table~\ref{tab:quant-summary-followup}).

\section{Discussion and Conclusion}
Across three seeds in one LLaDA+WikiText setting, our main replicated findings are that coarse linguistic labels are more linearly recoverable than exact lexical identity under the current probe setup, trajectory-level uncertainty tracks eventual correctness even late in denoising, and re-masking sensitivity peaks in the middle of the trajectory while remaining mostly local to the perturbed positions. Commitment provides additional evidence that denoising is temporally structured rather than uniform.

Denoising time is therefore a useful analysis axis for DLM interpretability. In future work, the same trajectory-first view could support stage-aware diagnostics, adaptive inference, or trajectory-level uncertainty estimation. Our claim, however, is intentionally narrow: we observe a coarse-to-fine pattern under these measurements, not a universal emergence law.

\section{Limitations}
We study one model family, one activation site, one masking regime, and one primary dataset. The lexical probe uses a compact label space with substantial unseen-target mass, so lexical comparisons should be read ordinally and alongside ranking metrics rather than as fully matched effect sizes. The 16-step, cross-dataset, and selector-restricted follow-ups are supportive but exploratory, and we do not claim that the same ordering must hold beyond the current setting.

%% file: sections/5-Appendix.tex
\section{Implementation and Experiment Details}
\begin{table}[t]
\centering
\small
\caption{Selected quantitative summaries from completed runs. Probe entries report step-0 to final accuracy; uncertainty entries report final confidence/entropy; 32-step values are three-seed means unless noted otherwise. The commitment-correctness entries use the representative seed-42 run.}
\label{tab:quant-summary-main}
\begin{tabular}{@{}p{0.28\columnwidth} p{0.32\columnwidth} p{0.28\columnwidth}@{}}
\toprule
Analysis & Statistic & Value \\
\midrule
Probe recoverability & POS ($0 \rightarrow \mathrm{final}$) & $57.9 \rightarrow 60.2$ \\
 & semantic ($0 \rightarrow \mathrm{final}$) & $59.8 \rightarrow 62.3$ \\
 & token ($0 \rightarrow \mathrm{final}$) & $13.4 \rightarrow 15.5$ \\
 & final POS-token gap & $44.7$ \\
 & WT2 final POS / sem. / token & $57.2 / 61.2 / 18.4$ \\
\addlinespace[2pt]
Commitment timing & NUM / NOUN / VERB & $3.37 / 4.07 / 4.46$ \\
 & FUNCTION / PUNCT & $4.92 / 4.97$ \\
\addlinespace[2pt]
Commitment and correctness & step 0 / steps 5--9 & $63.0 / 29.8$ \\
 & steps 10+ & $47.3$ \\
\addlinespace[2pt]
Uncertainty & correct conf. / ent. & $0.877 / 0.774$ \\
 & wrong conf. / ent. & $0.819 / 1.173$ \\
\bottomrule
\end{tabular}
\end{table}

\begin{table}[t]
\centering
\small
\caption{Perturbation robustness summaries from completed runs. Entries report the peak final-accuracy drop after re-masking and resuming denoising, together with the step at which that peak occurs. The 32-step baseline is averaged over three seeds; the ratio sweep uses the representative seed-42 run.}
\label{tab:quant-summary-perturb}
\begin{tabular}{@{}p{0.28\columnwidth} p{0.32\columnwidth} p{0.28\columnwidth}@{}}
\toprule
Setting & Statistic & Value \\
\midrule
32-step baseline & peak drop & $10.9$ \\
 & peak steps & $15\text{--}18$ \\
 & peak direct / collateral drop & $10.92 / 0.03$ \\
 & step-2 direct / collateral drop & $0.16 / 2.96$ \\
 & WT2 peak step / drop & $23 / 11.2$ \\
\addlinespace[2pt]
16-step schedule & peak step & $15$ \\
 & peak drop & $9.5$ \\
 & final POS / sem. / token & $63.9 / 66.0 / 15.1$ \\
\addlinespace[2pt]
10/20/40\% re-masking & peak drops & $7.1 / 11.2 / 19.9$ \\
 & peak steps & $17\text{--}18$ \\
\addlinespace[2pt]
Targeted re-masking & committed peak step / drop & $15,20 / 11.5,10.5$ \\
 & uncommitted peak step / drop & $21,20 / 11.1,10.6$ \\
 & content / function peak step / drop & $18 / 10.5,\; 27 / 10.9$ \\
\addlinespace[2pt]
Initial mask ratio & 0.3 final POS / sem. / token & $59.6 / 62.2 / 12.5$ \\
 & 0.3 peak step / drop & $10 / 12.0$ \\
 & 0.5 final POS / sem. / token & $48.7 / 55.2 / 11.5$ \\
 & 0.5 peak step / drop & $31 / 8.5$ \\
\bottomrule
\end{tabular}
\end{table}

\begin{figure}[t]
  \centering
  \includegraphics[width=\linewidth]{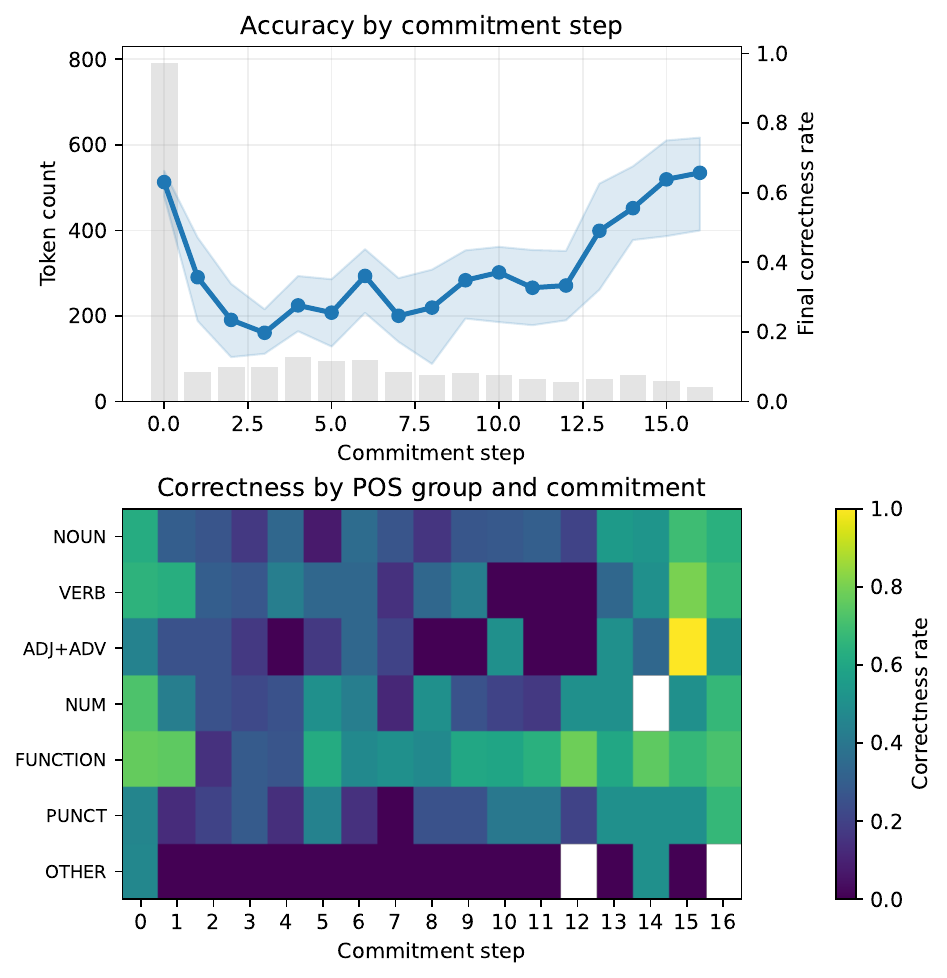}
  \caption{Commitment time is not a monotonic proxy for correctness: step-0 commitments are often right, mid-trajectory commitments are much less reliable, and very late commitments partially recover, with clear POS-dependent differences.}
  \label{fig:commitment-correctness}
\end{figure}

\paragraph{Experimental protocol.}
All main-text experiments use \texttt{GSAI-ML/LLaDA-8B-Base} on masked \texttt{Salesforce/wikitext} (\texttt{wikitext-103-raw-v1}) text with 32 denoising steps and final-layer activations recorded for masked positions only. We execute three independent seeds (42, 43, and 44). For each run, the dataset builder samples 1{,}000 probe-training sequences and 200 held-out analysis sequences, each 10--25 tokens long with a 40\% initial mask ratio, yielding 1{,}873 masked token positions on the evaluation split per run. Panels (b) and (d) of Figure~\ref{fig:main} report stepwise mean $\pm$ one standard deviation over these three runs. Panels (a) and (c), together with Figure~\ref{fig:commitment-correctness}, show the seed-42 run as a representative trajectory because the corresponding qualitative patterns are visually similar across seeds and the key replicated statistics are reported in text.

\paragraph{Shared probe training.}
All probe families use a shared linear decoder across timesteps rather than fitting one probe per step. For each training record, we concatenate hidden states from all denoising steps and train a single linear classifier on the resulting pooled examples. In the current setup, the optimizer is AdamW with learning rate $10^{-3}$, weight decay $0$, and one training epoch; we do not use class weighting or early stopping.

\paragraph{Labeling pipeline.}
POS labels come from a public BERT POS tagger and are mapped into coarse groups through the explicit \texttt{MODEL\_TO\_UNIVERSAL\_POS} and \texttt{UNIVERSAL\_TO\_COARSE\_POS} tables used by the analysis code. The semantic-category probe combines word-level predictions from a public BERT NER model with rule-based remapping into six labels: \textsc{Entity}, \textsc{Number}, \textsc{Content}, \textsc{Function}, \textsc{Punct}, and \textsc{Other}. This probe should be read as a coarse lexical-semantic/token-type surrogate rather than as a full semantic ontology.

\paragraph{Confidence, entropy, and token labels.}
At every denoising step, confidence is the top-1 probability under the full-vocabulary softmax, and entropy is the corresponding full-vocabulary softmax entropy. The token-identity probe does not classify over the entire vocabulary; instead, it uses the compact set of masked target IDs observed in the saved trajectory table for a run, yielding 3{,}428 classes on average across the three 32-step seeds. Relative to this compact task, uniform chance is 0.029\%, the train-majority baseline is 3.76\%, and 33.4\% of evaluation masked tokens belong to targets absent from the corresponding probe-train split. For that reason, comparisons to POS and semantic probes are interpreted ordinally rather than as normalized effect sizes. Final top-5 and top-10 retrieval nevertheless reach 29.4\% and 34.4\%, with final mean reciprocal rank 0.218, which makes the low top-1 scores substantially easier to interpret.

\paragraph{Perturbation procedure.}
For perturbation at step $t$, we first identify masked positions that have already been filled by the trajectory up to step $t$ and re-mask a random subset of those positions. If no masked position has been filled yet, we fall back to sampling from all masked positions. Randomness is controlled deterministically by the global seed together with the record and step indices, so repeated perturbation runs are reproducible.

\begin{table}[t]
\centering
\small
\caption{Offline follow-up summaries from completed analyses. Token-probe and probe-ablation entries report compact label-space statistics and local re-analyses of the three 32-step seeds. Confidence-interval entries report 95\% bootstrap intervals. Calibration values are from the representative seed-42 run.}
\label{tab:quant-summary-followup}
\begin{tabular}{@{}p{0.28\columnwidth} p{0.32\columnwidth} p{0.28\columnwidth}@{}}
\toprule
Analysis & Statistic & Value \\
\midrule
Token probe task & compact classes & $3428$ \\
 & uniform / majority baseline & $0.029 / 3.76$ \\
 & unseen eval-token fraction & $33.4$ \\
 & overall final top-1 / top-5 & $15.5 / 29.4$ \\
 & overall final top-10 / MRR & $34.4 / 0.218$ \\
 & seen final top-1 / top-5 & $23.4 / 44.1$ \\
 & seen final top-10 / MRR & $51.7 / 0.327$ \\
 & unseen final top-1 / top-5 & $0.0 / 0.0$ \\
 & unseen final top-10 / MRR & $0.0 / 0.001$ \\
\addlinespace[2pt]
Probe ablation & shared final POS / token & $60.2 / 15.5$ \\
 & per-step final POS / token & $58.6 / 14.6$ \\
 & per-step min POS-token gap & $34.5$ \\
 & shared / per-step token best step & $16 / 0$ \\
\addlinespace[2pt]
Bootstrap support & final POS-token gap & $44.7\;[43.1, 46.4]$ \\
 & final semantic-token gap & $46.8\;[45.2, 48.5]$ \\
 & 32-step peak drop & $10.4\;[9.9, 11.2]$ \\
\addlinespace[2pt]
Commitment strata & early / mid / late acc. & $62.8 / 35.1 / 33.8$ \\
 & early / mid / late peak drop & $12.8 / 11.1 / 12.9$ \\
\addlinespace[2pt]
Calibration drift & ECE step 0 / final & $0.034 / 0.415$ \\
 & Brier step 0 / final & $0.126 / 0.414$ \\
\bottomrule
\end{tabular}
\end{table}